# Bias, Fairness, and Accountability with AI and ML Algorithms

Nengfeng Zhou, Zach Zhang, Vijayan N. Nair,
Harsh Singhal, Jie Chen, and Agus Sudjianto
Corporate Model Risk, Wells Fargo

5/6/2021

*Abstract*

The advent of AI and ML algorithms has led to opportunities as well as challenges. In this paper, we provide an overview of bias and fairness issues that arise with the use of ML algorithms. We describe the types and sources of data bias, and discuss the nature of algorithmic unfairness. This is followed by a review of fairness metrics in the literature, discussion of their limitations, and a description of de-biasing (or mitigation) techniques in the model life cycle.

## 1. Introduction

Artificial intelligence (AI) techniques are used increasingly in many areas of applications, including banking and finance. They have several advantages over traditional statistical methods: i) ability to handle new data types such as text, audio, and images; ii) flexible models that yield excellent predictive performance; and iii) ability to automate many of the routine, and time-consuming, tasks in model development. However, the use of these algorithms also raise several challenges. A well-known problem is the opaqueness of ML models and the difficulties in understanding and interpreting the model results. In this paper, we focus on a related and equally important challenge: potential for bias and lack of fairness when using AI/ML techniques. This is currently a "hot topic" with increasing number of publications, conferences, and discussions. While the definitions of discrimination and fairness depend on the particular application, the most relevant ones for our purpose are those associated with protected groups.

There are regulatory requirements in banking aimed at preventing discrimination. For example, the Fair Housing Act (FHA, 1968) and Equal Credit Opportunity Act (ECOA, 2017) prohibit unfair and discriminatory practices based on protected attributes. ECOA explicitly mentions nine categories: race, color, religion, sex, national origin, age, marital status, receipt of public assistance, or any right exercised under the Consumer Credit Protection Act. Financial institutions can be liable for the following:

a) Disparate treatment: Differential treatment of members of a protected group compared to others, after taking into account relevant factors in the decision process. Disparate treatment usually happens in decisions based on judgement where the protected variable may be used explicitly or implicitly by the human decision makers; and

b) Disparate impact: Differential (or adverse) impact of a seemingly neutral decision, or policy, on members of a protected group compared to others. Disparate impact is mainly a concern in model-based decision-making. Even if protected variables are not used in a model directly, there is a possibility that other variables may serve as their proxies. Note, however, that the Consumer Financial Protection Bureau (Klein, 2019) has an important qualification to this rule: disparate impact will not create a violation if the necessity can be justified and no alternative decision or policy can have comparable performance with less discriminatory effect.

Traditional statistical methods have been used to test for disparate treatment in redlining cases and for checking consistency in pricing and underwriting. Testing for disparate impact, however, is a multistep process that involves determining if a protected class has been adversely affected, then looking for a justification for that specific policy or practice, and finally searching any alternatives that would result in less impact.

Fairness issues arise in banking and finance, criminal justice, social programs, healthcare, recruiting, marketing, and so on. One study that has received national attention deals with recidivism of criminal defendants (Larson, Mattu, Kirchner, & Angwin, 2016) . An analysis of the popular COMPAS algorithm (Correctional Offender Management Profiling for Alternative Sanctions) found that there was *a higher false positive rate in identifying black defendants to be at risk of recidivism* compared to white defendants; conversely, there were *lower false negatives for flagging whites as low risk* compared to blacks**.** The article at the above link cites several other examples of discrimination in social/criminal justice. See also the draft book (Barocas, Hardt, & Narayanan, 2020) for examples as well as state-of-the-art research discussion in this area. (Hutchinson & Mitchell, 2019) provides a good discussion of issues within education and hiring. One example that specifically documents fairness issue with ML algorithms (bias against women) is the Amazon's recruiting tool. See also the recent book 'The Ethical Algorithm' (Kearns & Roth, 2019) for other examples and easy-to-read discussion of research developments.

Within the banking industry, AI/ML techniques are being used in consumer lending (credit scoring and more recently marketing and collections), conduct analysis and compliance management. Fair lending considerations are at the forefront in credit scoring and decisioning. Conduct analysis and compliance management are emerging areas that also have significant potential for bias and unfairness. They rely on analysis of unstructured data (text, e-mails, etc.) using natural language processing (NLP) and ML techniques.

The rest of the paper is organized as follows. Section 2 provides a brief overview on AI/ML algorithms and specifies those of interest to the discussion in this paper. This is followed by a description of the sources and types of bias and fairness issues (Section 3). Section 4 provides a review of fairness metrics in the literature and their limitations. Section 5 discusses de-biasing (or mitigation) approaches and how they can be used at different stages of the model lifecycle. Section 6 outlines some general guidelines on ensuring fairness.

## 2. Scope of AI/ML Algorithms

While the phrases *Artificial Intelligence* and *Machine Learning* are often used interchangeably in popular discussion, they are quite different. AI is much broader in scope and ML is just one of the pathways to accomplishing the goals of an AI project. AI has a very long history dating back to formal reasoning in logic, philosophy and other fields. The term itself was coined by John McCarthy only in 1959 as:

- *The study of "intelligent agents" – devices that perceive the environment and take actions that maximize its chance of success at some goal.*

AI has had mixed successes in the past and has gone through periods of "AI winters". There has been a massive resurgence recently, due to the amounts of data available and exponential leaps in computing power. In addition, the development of deep learning neural networks with their excellent predictive performance, especially for pattern recognition, has led to much excitement.

The term *Machine Learning* was proposed by Arthur Samuel in 1959 who described it as:

- *"A field of study that gives computers the ability to learn without being explicitly programmed."*

A more engineering-oriented definition was given by Tom Mitchell in 1997:

- *A computer program is said to learn from experience E with respect to some task T and some performance measure P, if its performance on T, as measured by P, improves with experience E.*

See (Hu L. , Chen, Nair, & Sudjianto, 2018) for references. It was not until the last two to three decades that there has been wide usage of ML techniques. The main reasons again are availability of massive amounts of data and advances in computing power, including capabilities in data storage, data transfer, data architecture, and fast computing.

Some people use the term ML very broadly to include even traditional statistical methods such as linear regression and clustering, techniques that have been around for a long time. In this paper, we will restrict usage to modern approaches for supervised, unsupervised, and reinforcement learning, including support vector machines, ensemble algorithms (random forest, gradient boosting), and various types of neural networks (feedforward, CNN, RNN, LSTM, GAN, Auto-encoders, Isolation Forest and so on).

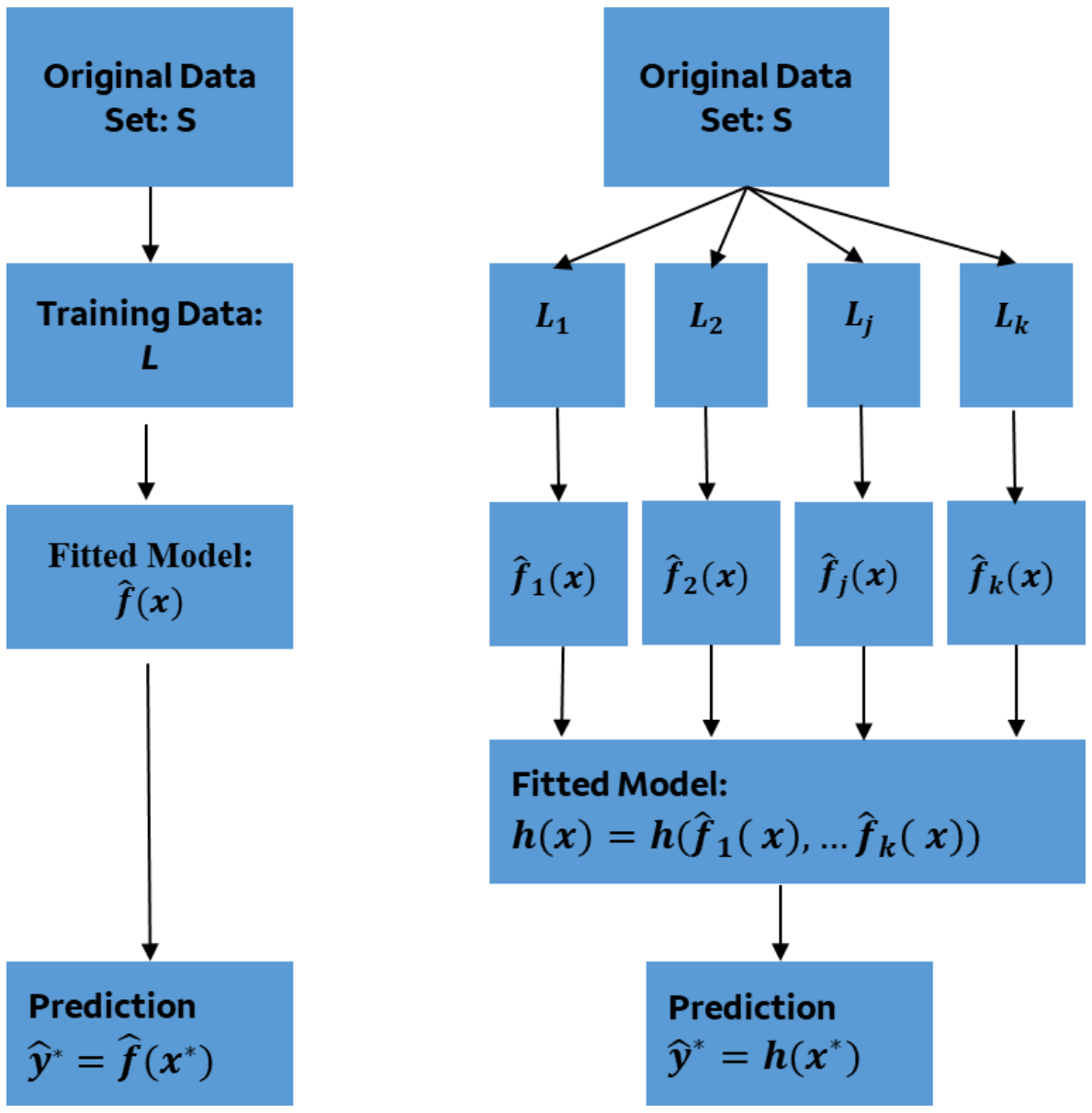

*Figure 1: A comparison of traditional and ensemble algorithms for supervised learning*

Figure 1 provides a visual comparison of traditional and ensemble methods for supervised learning (see Hu, Chen, Nair, & Sudjianto, 2018). $\mathbf{S}$ refers to the original dataset, $\mathrm{L_k}$'s refers to the training datasets, and $\hat{f}_k$'s refer to the fitted models on the training datasets. In the traditional framework, we fit a single regression model and the results are easy to interpret. In ensemble models such as Random Forest (RF) and Gradient Boosting (GBM), the models are fitted across multiple training datasets and are aggregated to get an overall model. See (Hu L. , Chen, Nair, & Sudjianto, 2018) for a more detailed tutorial. While the overall models are very flexible and generally have very good predictive performance, they are also complex and difficult to understand/interpret.

Figure 2 provides an example of a simple feedforward neural network (NN) with one hidden layer and multiple (three) nodes. The various inputs (hand-crafter predictors or original observations such as time series or images) are combined at the hidden layer. An activation function is then used to filter these values, and the results are then further combined to get the final output. In practice, there are multiple layers and more complex network architecture (such as convolutional and recurrent neural networks).

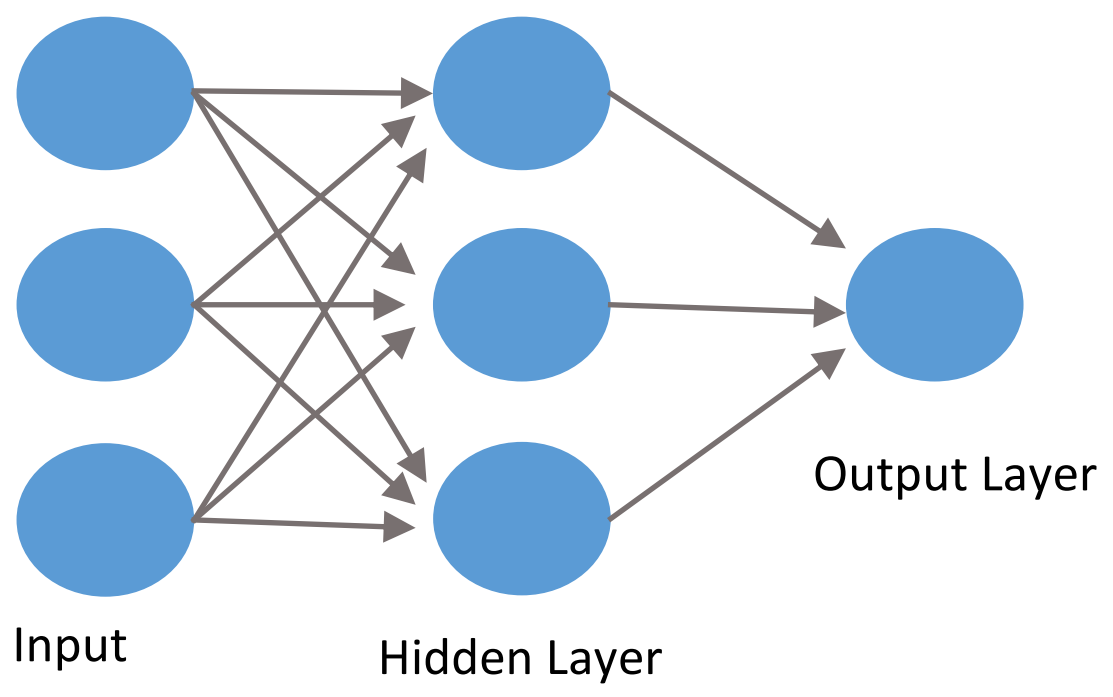

*Figure 2: A simple feedforward neural network with a single hidden layer and multiple nodes*

Ensemble algorithms and neural networks are also used in unsupervised applications. Examples include isolation forests for anomaly detection, auto-encoders for non-linear dimension reduction, and generative adversarial networks for simulating additional data from a given sample.

The main takeaway from this discussion is that these ML models are very flexible and have good performance but are complex. The results are hard to understand and interpret, and thus have consequences for bias and fairness.

## 2. Potential Sources of Bias and Discrimination

Fairness considerations have been around for a long time. For example, (Courchane, Nebhut, & Nickerson, 2000) describe the results of a case study on fair-lending examinations of national banks from 1994-1999 and note that despite "years of intense scrutiny, lending discrimination still persists." The arrival of flexible and automated AI/ML algorithms as well as the availability of alternative sources of data are leading to new challenges and exacerbating current ones.

We group the potential sources of bias and fairness issues into two broad areas – data bias and algorithmic bias – and discuss each of them below:

## 3.1 Data Bias

a) Bias in historical data: Historical data are often skewed towards, or against, particular groups. Data can also be severely imbalanced with limited information on protected groups. The literature is full of examples where historical biases have resulted in discrimination. Recall the COMPAS study mentioned earlier. Biases in historical data can be exacerbated with the use of ML techniques for the following reasons. Users tend to include a large numbers of input variables in the model because ML algorithms can automate feature engineering. Further, time-consuming tasks such as checking for multi-collinearity and variable selection are not needed in ML. As a result, it is easy to ignore biases in the data and also ignore the presence of surrogate variables for protected attributes. Recall the earlier example on Amazon's recruiting tool that resulted in bias against women because the data was biased towards males who dominated the technology industry.

b) Bias in data collection mechanisms: Advances in data capture technologies have made it easier to collect different types of data. But insufficient attention is being paid to inherent biases in the data collection mechanisms and lack of representativeness. A well-known example is "crowdsourcing" of pothole information where smart phones are used to automatically sense and transmit data when a vehicle goes over potholes. This mechanism is clearly biased towards the segment of the population that owns smart phones and the roads they use. There are many other examples in the literature, including development of marketing campaigns based on the usage of the internet, and use of video imaging and facial recognition systems that have better performance for people with lighter skin tones and have higher error rates in other cases.

c) Bias in alternate sources of data: Much of the excitement with the big data phenomenon is due to new sources of readily available data: worldwide web, social media, blogs, etc. A recent paper (Berg, Burg, Gombović, & Puri, 2018) demonstrates problems with such alternate sources of data in the context of credit scoring. They show that "digital footprint" variables, such as borrower's computer (Mac vs PC), type of device (phone, tablet, PC), whether name is part of the borrower's e-mail, have strong predictive performance in credit scoring. However, these predictors are highly correlated with socio-economic variables that are surrogates for protected groups. Another example discussed in (Klein, 2019) is data on divorce proceedings that are good predictors of potential bankruptcy. More often than not, these variables are highly related to protected classes, and the availability of such seemingly innocuous information, combined with flexible "data snooping" ML algorithms, can easily lead to "proxy discrimination".

d) Unobservable Outcomes: (Corbett-Davies & Goel, The Measure and Mismeasure of Fairness: A Critical Review of Fair Machine Learning, 2018) discuss measurement problems (bias) associated with labels/responses and predictors. Unobservable outcomes (or labels) is a huge issue, one that makes it difficult to even measure discrimination. Consider an example with mortgage applications where the lender uses a model to decide on approving or denying the loan. In term of "true" outcomes, we observe loan defaults only for those who received a mortgage and we do not have any information for those who were denied mortgage – whether they would have defaulted or not had they been approved for a mortgage (a counterfactual). Thus, we have true outcomes (default or not) for only part of the population. (Reject inference is a technique used in banking to try to address this problem.) The same issue arises in job applications when we use a model or even interviews to decide on job offers. True outcomes (job performance that measure whether candidates are qualified or not) are available only for those who actually got the job. In the absence of information on all outcomes, one cannot even measure fairness.

e) Bias in unstructured data and feature engineering: Unstructured data, such as texts, audio, and images, are increasingly analyzed through AI/ML techniques in banking and finance. In these applications, one converts data summaries and selected features, such as word embeddings, for analysis. However, the feature selection process can suffer from "implicit biases". Figure 3 shows word embeddings that suffer from gender bias. A commonly quoted example of a word embedding is "father is to doctor as mother is to nurse".

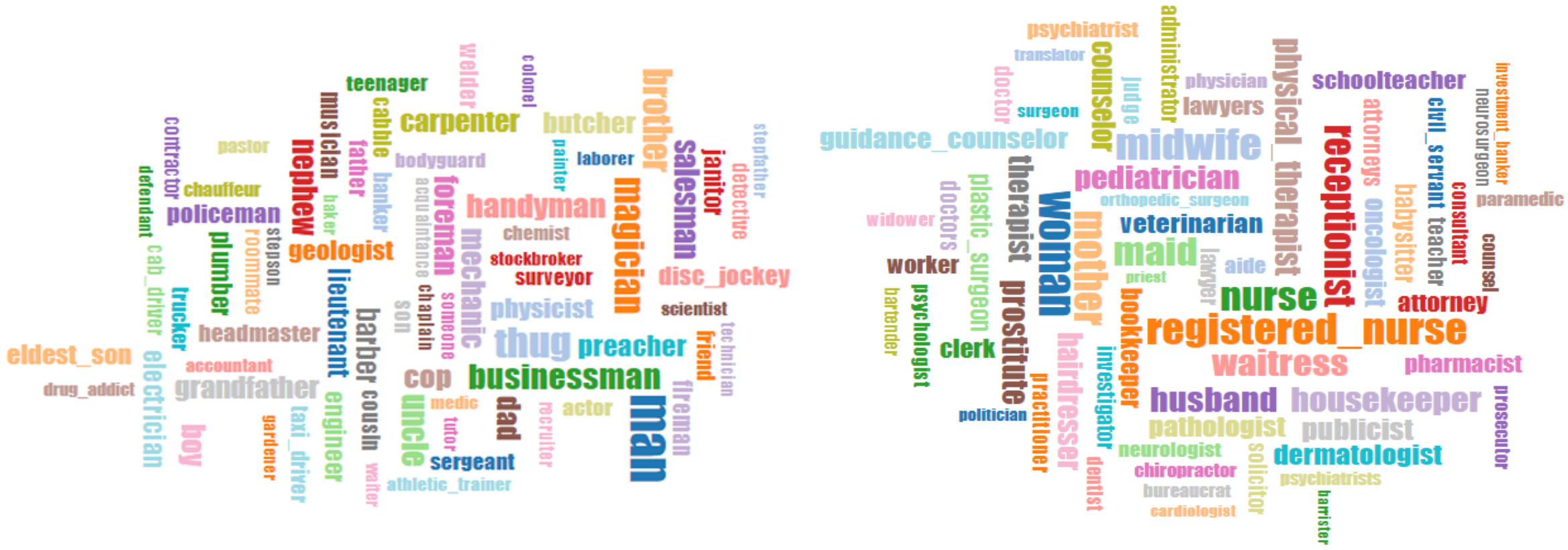

*Figure 3: Example of gender bias in word embedding* (Buonocore, 2019)

## 3.2 Algorithmic Bias

f) The automated nature of modern ML algorithms present its own challenge. Model developers in banks are using data at account levels (millions of observations) with thousands of predictors. In automated use of ML, the algorithms artificially create hundreds of derived predictors (by transforming the original ones) with the hope of getting tiny improvements in predictive performance. However, such processes do not incorporate subject-matter knowledge to carefully review the selected variables and miss the potential for correlated surrogate variables causing proxy discrimination.

g) The flexible nature of the algorithms can lead to overfitting of training data. In fact, one could ask whether the increased predictive performance is in fact good model fitting or is it good memory. The algorithms can "recognize" the data and even individual observations, leading to concerns of privacy as well as fairness. There are techniques for addressing the over-fitting problem including the use of test/validation data and regularization methods. However, these approaches do not fully address the problem, especially if the test/validation datasets also have the same kinds of biases present in the training dataset. An interesting anecdote relates to how *Baidu* tried to get a 'leg up' in a Kaggle competition by using several e-mail accounts to get multiple views of the validation dataset (https://www.technologyreview.com/s/538111/why-and-how-baidu-cheated-an-artificial-intelligence-test/).

h) Optimization of ML algorithms can also lead to biases. The primary objective in hyperparameter tuning and optimization is maximizing predictive performance. In doing so, one ignores other important issues such as robustness, fairness, etc. Unlike traditional parametric models, ML algorithms are so flexible that they can learn every bit of information in the data and end up amplifying small biases in the data. See (Hooker, 2021).

i) Data bias together with poor optimization of algorithms can cause severe harm to protected groups, especially when information on such groups is under-represented in the training data. See also (Hooker, 2021). In such situations, the results from ML algorithms can be inaccurate or the model can be very unstable.

j) A related concern is the opaqueness and lack of interpretability of complex ML algorithms. For traditional models, which are typically global, one can measure fairness in a global fashion. Many tree-based ML algorithms, such as RF and GBM, are inherently local, and a different approach is needed to measure fairness locally. If one can identify the input-output relationships, including complex interactions and local behavior, one can isolate potential algorithmic bias. There are techniques available for exploring such relationships in the context of ML algorithms relationships (see Chen et al. 2020). However, these tools are currently limited in their ability to tell the full story. Thus, one may not be able to identify the presence or reasons for discrimination.

## 4. Fairness Metrics

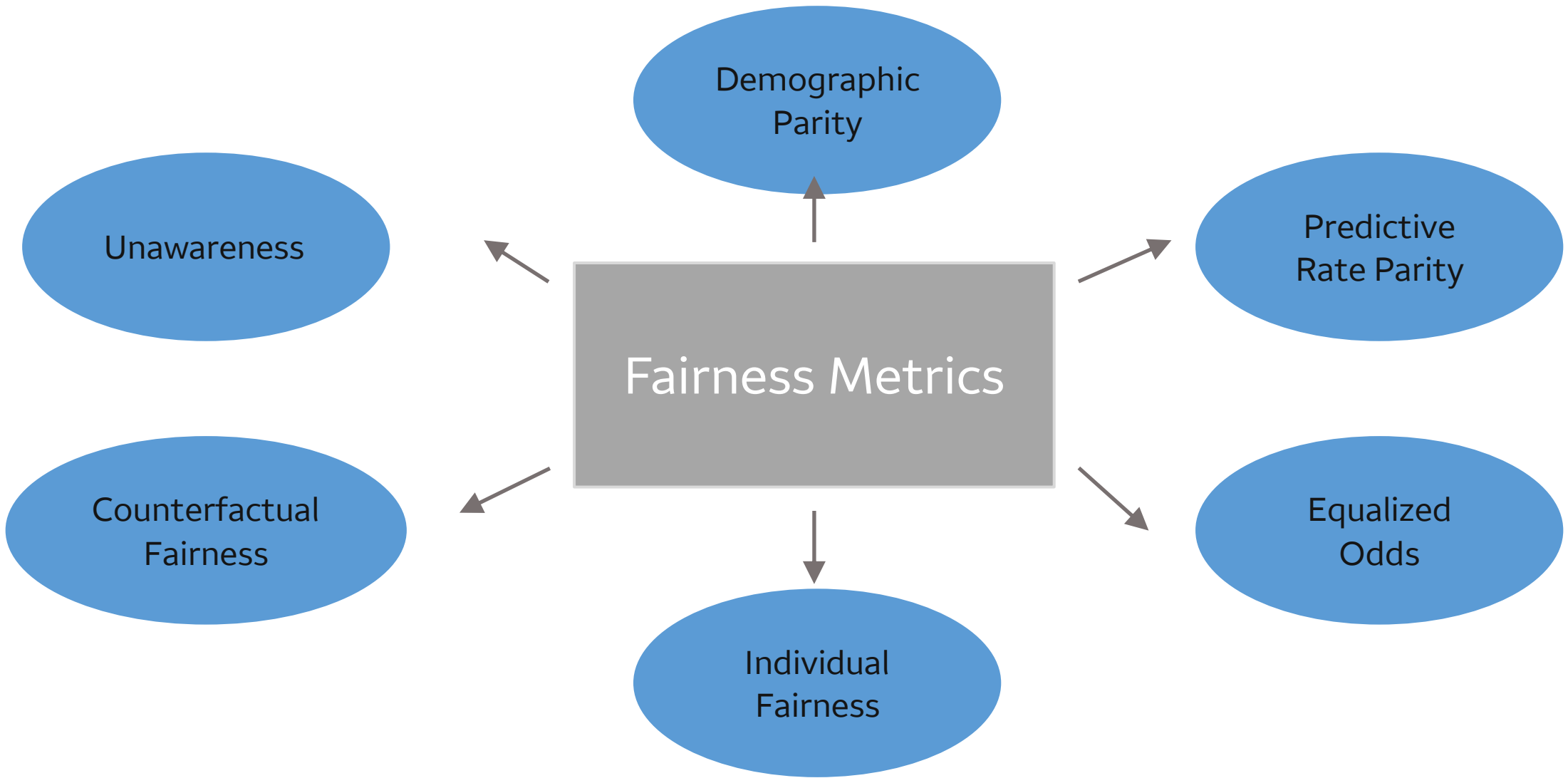

*Figure 4: Different fairness metrics*

An important problem in bias and fairness discussions is how to define fairness. There is no universally accepted definition, and researchers have tried to approach the problem from different angles. Figure 4 shows common and widely used metrics to assess model or algorithmic fairness. We discuss them in detail below. See also (Mitchel, Potash, Barocas, D'Amour, & Lum, 2019) for a survey of different categories "fairness" related to ML and statistical model-based predictions.

**a) Group Fairness**

We start with a definition of relevant notation and use the recidivism example for illustration.

- $Y \in \{0, 1\}$: binary response variable (whether a prisoner will recidivate or not after he is released);
- $\hat{Y} \in \{0, 1\}$: binary prediction variable (result of an algorithm used by, say, a parole board, to predict recidivism);
- $X$: $p$ −dimensional feature associated with the defendant, e.g. education, work experience, past criminal history, etc.); and
- $A \in \{0, 1\}$:a binary protected attribute, e.g. gender (in general, it can be multi-dimensional and take on multiple values).

(Note: In this recidivism example, if the parole board used the prediction on recidivism to decide on parole, it might in fact end up affecting the true recidivism outcome. We assume that this is not the case here, but this can a problem in practice.)

Here are some common group fairness metrics:

i. Demographic parity aims to ensure that probability of recidivating is equal across the sensitive attribute:

$$P\left[\hat{Y} = j \middle| A = 0\right] = P\left[\hat{Y} = j \middle| A = 1\right], j \in \{0, 1\}.$$

In our recidivism example, this means that the predicted recidivism rates for males and females must be equal.

This metric has been commonly used in different areas, and there are variations that enforce it subject to constraints such as$P\left[\hat{Y} = 1 \middle| A = 1\right] \leq const \times P\left[\hat{Y} = 1 \middle| A = 0\right]$). (Hardt, Price, & Srebro, 2016) noted that, while this metric is "simple and seemingly intuitive, it has many conceptual problems". See also (Dwork, Hardt, Pitassi, Reingold, & Zemel., 2012). For instance, women are known to have lower recidivism rates compared to similarly situated men. By enforcing demographic parity, we may be treating women unfairly.

ii. Predictive rate parity metric means the predictive value should be equal for the protected and unprotected class (Verma & Rubin, 2018):

$$P[Y = k|\hat{Y} = j, A = 0] = P[Y = k|\hat{Y} = j, A = 1], k, j \in \{0, 1\}$$

In other words, $Y$ is independent of $A$ conditional on $\hat{Y}$. In the recidivism example, this metric implies that, conditional on predicted recidivism, the true recidivism rate must be equal for men and women.

iii. Equalized odds requires that the false positive rates and false negative rates should be equal for the protected and unprotected class (Zhong, 2018).

$$P\left[\hat{Y} = j \,\middle|\, Y = k, A = 0\right] = P\left[\hat{Y} = j \,\middle|\, Y = k, A = 1\right], k, j \in \{0,1\}$$

In our example, for the algorithm to be fair, the predicted recidivism probabilities must be the same for men and women across all classes.

iv. Equal opportunity is a special case of equalized odds where the equality is satisfied only for the special case $Y = j$ , where *j* is the "advantageous" outcome.

$$P\left[\hat{Y} = j \,\middle|\, Y = j, A = 0\right] = P\left[\hat{Y} = j \,\middle|\, Y = j, A = 1\right]$$

This is a weaker condition than equalized odds where the equality is enforced only for the "advantageous" outcome.

v. Conditional parity measures fairness conditional on the values of some appropriate variables (Ritov, Sun, & Zhao, 2017). For example, in fair lending, it is reasonable to assess fairness conditionally on legitimate credit characteristics such as FICO. To define the metric, let *U* denote information that is being conditioned on. For illustrative purpose, assume U takes on discrete values. Then, we have

$$P\left[\hat{Y} = j \,\middle|\, U = k, A = 0\right] = P\left[\hat{Y} = j \,\middle|\, U = k, A = 1\right], k \in \{0,1,..K\}, j \in \{0,1\}.$$

See also (Zhang, Lemoine, & Mitchell, 2018) for a comprehensive review of these metrics.

**b) Individual fairness**

This concept states that "similarly situated individuals" should be treated similarly (Verma & Rubin, 2018). One way of enforcing this is to: i) define a suitable distance metric on features associated with individuals and on the decision space; and ii) ensure that if the distance between two individuals is small, the corresponding distance between their decision distributions is small. Of course, the challenge is to identify meaningful and non-controversial distance metrics.

**c) Unawareness (anti-discrimination)**

This refers to situations where the protected attribute(s) are explicitly removed from the data before the model is trained, so the model is "not aware" of the sensitive attribute(s). This metric is easy to implement and use. But the obvious limitation is that information in sensitive attributes is often present in surrogate variables (proxy discrimination). For example, it is known that zip codes could serves as proxies for race. This problem gets particularly acute with large datasets (and alternate datasets) where such information is present but hard to detect due to the use of automated algorithms for variable selection. For this and other reasons, some researchers in the literature have argued that it is better to collect and appropriately use information of protected attributes.

**d) Counterfactual Fairness**

This metric is related to the area of counterfactual modeling that arises in causal inference. Counterfactual fairness was proposed in (Russell, Kusner, Loftus, & Silva, 2017). In this case, one would consider a counterfactual scenario where (in our recidivism example) the defendant had a different protected attribute – say female instead of male. If we would have made the same decision regardless of gender, then the decision is considered to be fair. Since this approach is not as commonly used as others, we do not consider it further.

**Limitations of Fairness Metrics:**

The concept of demographic parity has been around for a long time. This, together with equalized odds and predictive rate parity, have been discussed by many authors. Interestingly enough, these metrics are in conflict with each other, and there is an "impossibility theorem of fairness" that states that any two of the three criteria are mutually exclusive. See (Chouldechova, 2017) and (Kleinberg, Mullainathan, & Raghavan, 2016). (Sam Corbett-Davies, 2018) present the limitations of three popular fairness metrics. They show that requiring unawareness (anti-classification) or equalized odds (classification parity) can hurt the protected groups by relabeling the model predictions to achieve the fairness; and predictive parity (calibration) is usually not strong enough to guarantee equity.

(Hardt, Price, & Srebro, 2016) describe a case study for FICO model where they measure the effectiveness of different fairness metrics. They found that equalized odds are very difficult to achieve in consumer lending even for the FICO model. It requires both the loan approval rate of non-defaulters and the loan approval rate of defaulters to be constant across groups. This cannot be achieved with a single threshold for each group, but requires randomization of prediction between two thresholds.

Fairness for some decisions are better decided at a group level than at individual levels. For example, in university admissions, many people have argued that a diverse student body will benefit the entire institution.

## 5. De-biasing and Mitigating Unfairness

Researchers have suggested a number of approaches and techniques for "de-biasing" or finding ways to mitigate unfairness. Some of these are drastic approaches that suggest, for instance, actively manipulating the data. Their potential is limited due to legal and compliance considerations. Nevertheless, we provide a review of the methods in this section.

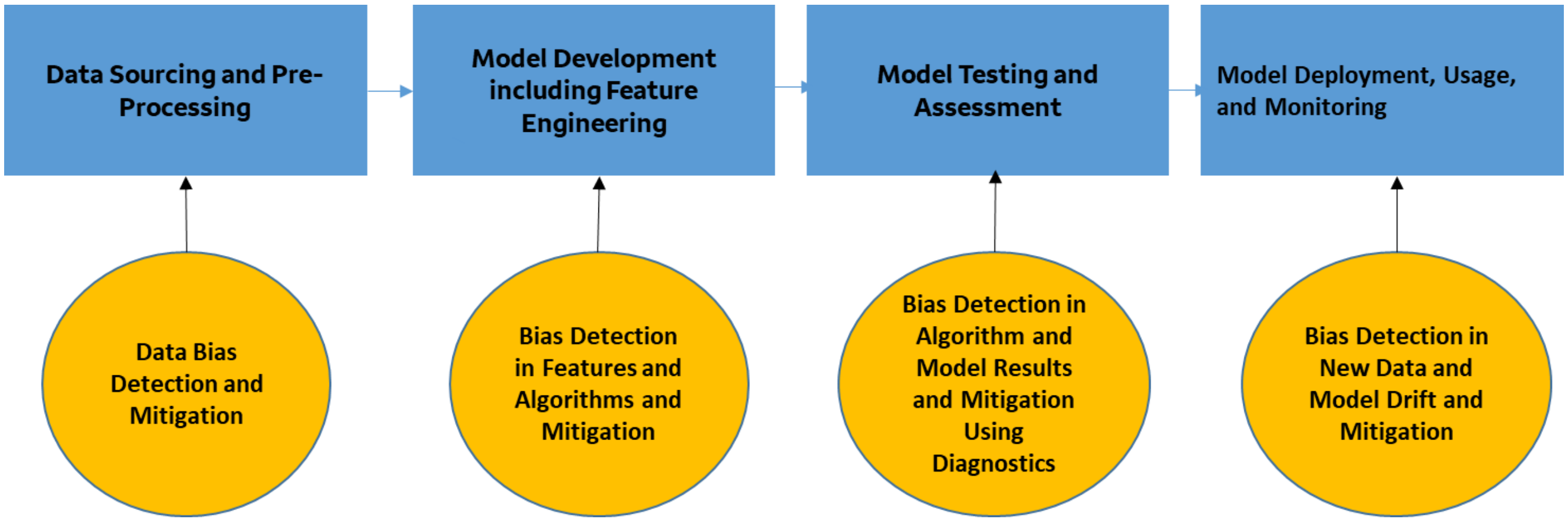

*Figure 5: An overview model life cycle, different types of biases and mitigation efforts*

Figure 5 shows an overview of the model lifecycle together with corresponding points for bias detection and mitigation efforts. The main stages in a model lifecycle are: i) data sourcing and pre-processing, ii) model development which includes variable selection, feature engineering, and model training; iii) model testing and assessment which includes assessing model fit, use of model diagnostics including model interpretation and explainability; and finally iv) model deployment, usage, and monitoring. One should develop and implement strategies to detect and reduce bias at the different stages of the model life cycle. Further, this should be done before the model is deployed and used in production.

Obviously, any bias mitigation effort should start with the data sourcing and pre-processing stage. Some of the techniques suggested in the literature include suppression of sensitive variables and their proxies, data massaging (changing outcome labels to correct data bias), and reweighting/sampling to promote certain instances while demoting others (Kamiran & Calders, 2012). A more formal approach, proposed in (Calmon, Wei, Vinzamuri, Ramamurthy, & Varshney, 2017), involves creating new data points that preserve the distribution of the original data as much as possible and mapping the original data to the newly-created data points to preserve demographic parity. This idea covers the model development stage as well since one cannot compute demographic parity without model results.

The above proposals are all fraught with danger and can lead to serious abuses. They are also likely to violate legal and compliance considerations. One should not engage in data massaging, suppression, or reweighting (as described above) without clearly understanding potential downstream consequences. Unfortunately, it is difficult to accomplish that with large datasets and complex machine-learning algorithms. The proposal in (Calmon, Wei, Vinzamuri, Ramamurthy, & Varshney, 2017) to create and use artificial data is even more subject to potential abuse. In our view, there is no good substitute for carefully reviewing the datasets for historical and measurement biases as well as checking for potential limitations and non-representativeness in data collection mechanisms. These challenges are magnified with large datasets and new alternate sources of data. Further, they go against the spirit of automation promised by AI/ML algorithms.

The second stage for bias detection and mitigation efforts is the 'in-processing' or "model development/training" stage. One natural approach here is to incorporate the fairness metrics in the estimation process as constraints. This can be a hard constraint where one optimizes the decision subject to a suitable fairness metric, say demographic parity. For example, we can optimize credit decisions has subject to the given threshold:

$$P[\hat{Y} = j | A = 1] \geq \mathbf{0.8}\ P[\hat{Y} = j | A = 0]$$

Alternatively, we can incorporate it as a soft constraint through regularization such as:

$$Min \{\sum_i L(y_i, \hat{y}_i) + \lambda \left(P[\hat{Y} = 1 | A = 1] - P[\hat{Y} = 1 | A = 0]\right)^2,$$

for suitable loss function $L$ and tuning constant $\lambda$. This is the approach taken in several papers in the literature. (Kamishima, Akaho, Asoh, & Sakuma, 2012) proposed ensuring demographic parity by adding a regularizer in the logistic regression setting. See also (Zemel, Wu, Swersky, Pitassi, & Dwork, 2013). (Zhang, Lemoine, & Mitchell, 2018) suggest a different approach based on an adversarial objective function along the lines of a GAN network. (Celis, Huang, Keswani, & Vishnoi, 2018) suggest a framework that uses a set of fairness metrics in a meta-algorithm. (Kamishima, Akaho, Asoh, & Sakuma, 2012) discuss ways to remove group prejudice using regularization. (Zemel, Wu, Swersky, Pitassi, & Dwork, 2013) consider the concept of individual fairness and address it together with group fairness. See also (Calders & Verwer, 2009), (Woodworth, Gunasekar, Ohannessian, & Srebro, 2017), (Zafar, Valera, Rodriguez, & Grummadi, 2017), (Agarwal, Beygelzimer, Dudik, Langford, & Wallach, 2018).

Recall however the earlier discussion on limitations and conflicts of the fairness metrics. Further, no single metric of fairness is uniformly better than other metrics. Therefore, the choice of fairness metric in this context has to be done carefully with the particular application as well as appropriate laws and regulation in mind. Note also that there is a trade-off between prediction accuracy and fairness. For instance, (Pleiss, Ragahavan, Wu, Kleinberg, & Weinberger, 2017) examined the trade-off between minimizing error disparity across different population groups while also maintaining calibrated probability estimates.

The next point for bias detection and mitigation efforts is the model testing and assessment stage. In addition to the traditional metrics for model performance, one should also compute the relevant fairness metrics and assess disparate treatment and impact. This may require refitting the model by incorporating fairness thresholds/constraints and refining any existing constraints. At an extreme, this may also require going back to the original data to identify and mitigate data bias. (Kamiran & Calders, 2012) discuss discrimination-aware classification that does not require data modification or modifying the classification results. This belongs to the "post-processing" stage. Note that the use of such algorithms require the protected attributes to be available in the model usage stage, which will introduce compliance risk.

Another important dimension at this stage is identifying sources of algorithmic bias. This would require interpreting and explaining the results from the complex algorithms. This has also been the subject of considerable research in academia and industry.

Finally, once a model has been deployed, it must be continually monitored for disparate impact testing in order to ensure that its predictive performance does not decline. Degradation in model performance can be caused by several factors: changes in the datasets, availability of new data, changes in environment that violates the model's assumptions (model drift). Any new types of bias in the model during production has to be compared to the development data. Model retraining may be needed to ensure that the model continues to be fair and accurate.

The Institute of International Finance survey (Bailey, 2019) addressed specific responses and solutions taken by financial institutions to ensure fairness. In addition to technical measures, they propose looking closely at the governance process for ML.

One important topic in fairness is the availability and usage of protected attributes. In our discussion of the fairness metrics calculations and de-biasing algorithm, we have assumed the protected attributes are available. However, except in mortgage loan applications, banks are not legally allowed to collect protected attribute data when a person is applying for a loa,. In assessing fair lending, the protected attributes (gender and race) are usually imputed from names and addresses. Such imputations can be inaccurate, especially for some racial groups. Thus, fairness cannot be assessed precisely and corrected. This will actually harm the group of people due to lack of protected attributes.

It is illegal in consumer banking industry and many other areas to use protected attribute for model training or decision. This may actually prevent the institutions in ensuring fairness since the de-biasing algorithm needs information on protected attributes. There are also situations where protected groups would get better treatment than non-protected groups if the protected attributes were used, such as the example of lower rates of recidivism for women.

## 6. Concluding Remarks

We close with a discussion of some key points to consider as we develop and implement AI/ML techniques in banking.

- Certain application areas, such as consumer lending, have potential for serious harm from use of black-box algorithms that are not well-understood. This view appears to be shared by regulators, and we expect application of AI/ML algorithms will be limited in these areas in the near future.

- Fairness concerns are heightened when alternative sources of data, such as social-media data, information on biometrics, speech or language, are used. In these cases, it is not easy to scrub the data of demographic proxies.

- There are multiple approaches to mitigating unfairness concerns. No single approach is universally best, and choosing the most appropriate one will require expert judgement as well as knowledge of relevant legal and compliance requirements.